\documentclass[fleqn,10pt]{wlscirep}
\usepackage[utf8]{inputenc}
\usepackage[T1]{fontenc}
\usepackage{algorithmic}
\usepackage{algorithm}
\usepackage{bm}

\title{Path Planning of Surgical Robot for Minimally Invasive Surgery on Riemannian Manifold}

\author[1]{Yoshiki Yamamoto}
\author[2,+]{Maina Sogabe}
\author[3]{Shunichi Hirahara}
\author[1]{Toshiki Kaisaki}
\author[4]{Tetsuro Miyazaki}
\author[4,*]{Kenji Kawashima}
\affil[1]{The University of Tokyo, Department of Information Physics and Computing,Tokyo, 113-8656, Japan}
\affil[2]{The University of Tokyo, Department of Precision Engineering, Tokyo, 113-8656, Japan}
\affil[3]{National Institute of Informatics, Principles of Informatics Research Division, Tokyo, 101-8430, Japan}
\affil[4]{The University of Tokyo,  Department of Information Physics and Computing, Tokyo, 113-8656, Japan}
\affil[*]{kkawa729@g.ecc.u-tokyo.ac.jp}
\affil[+]{these authors contributed equally to this work}


\begin{abstract}
Robotic surgery for minimally invasive surgery can reduce the surgeon's workload by autonomously guiding robotic forceps. 
Movement of the robot is restricted around a fixed insertion port.
The robot often encounters angle limitations during operation. 
Also, the surface of the abdominal cavity is non-concave, making it computationally expensive to find the desired path.
In this work, to solve these problems, 
we propose a method for path planning in joint space by transforming the position into
a Riemannian manifold. 
An edge cost function is defined to search for a desired path in the joint space and reduce the range of motion of the joints.
We found that the organ is mostly non-concave, making it easy to find the optimal path using gradient descent method.
Experimental results demonstrated that the proposed method reduces the range of joint angle movement compared to calculations in position space.

\end{abstract}
\begin{document}

\flushbottom
\maketitle
%
%
\thispagestyle{empty}


\section*{Introduction}
Minimally invasive surgery (MIS) has revolutionized operative medicine by reducing patient trauma, minimizing incision size, and accelerating postoperative recovery\cite{MIS}. 
The advantage of surgical robots has further enhanced MIS capabilities\cite{Mare}$^{,}$\cite{Emer}.
To date, over one million robotic procedures have been performed worldwide, yet control remains purely teleoperated, with the robot acting as a sophisticated input/output device under continuous surgeon guidance\cite{Kawa}$^{,}$\cite{Cepo}.
The global shortage of trained robotic surgeons,  motivates the development of semi‐ and fully‐autonomous functionalities\cite{kawa1}$^{,}$\cite{Tam}$^{,}$\cite{Han}. Key challenges include safe trajectory generation within patient‐specific anatomy, real‐time collision avoidance around vital structures.

Traditional motion planning for robots is formulated in position space, where the end‐effector pose is planned and then converted to joint commands via inverse kinematics. Sampling‐based planners such as Probabilistic Roadmaps (PRM) \cite{PRM}$^{,}$ \cite{PRM1}$^{,}$\cite{PRM2}$^{,}$\cite{
PRM3}and Rapidly‐Exploring Random Trees (RRT) \cite{RRT}$^{,}$\cite{kara}$^{,}$\cite{Wei}$^{,}$\cite{RRT2} are popular due to their probabilistic completeness in high‐dimensional spaces. 
Deep Reinforcement Learning (DRL) is a powerful tool to for path planning. DRL is apply trajectory planning of the robotic manipulator in complex environments\cite{DRL}.
A novel reinforcement learning framework that combines a modified artificial potential field was proposed \cite{DRL2}.
Another approach, configuration space planning, which samples directly in joint space, inherently respects mechanical constraints and port pivots. Configuration space PRM/RRT methods have been applied to articulated arms and industrial robots \cite{Li}$^{,}$\cite{Meng}$^{,}$\cite{Kang}, but explicit mapping of complex anatomical obstacles into joint space remains challenging. Early approaches used voxel or mesh approximations for obstacle projection, suffering from high memory demands or coarse collision checks. More recent efforts employ smooth barrier functions\cite{Chair}$^{,}$ \cite{BF2}or Riemann metrics to embed obstacle avoidance into continuous optimization\cite{Klein}. 
MRI images show that the surfaces of organs can be represented as a combination of convex surfaces, while the walls of body cavities are concave\cite{MRI}.
The gradient method often used in collision avoidance such as SDF (Signed Distance Field) methods \cite{SDF1}$^{,}$ \cite{SDF2}
can compute paths efficiently on non-concave surfaces. 

For surgical robots for MIS, we proposed a path planning method on Riemannian manifolds \cite{Rie} 
in joint space, instead of position space.
Contributions of this paper are as follows:
First, we propose a method for path planning in joint space by transforming position space into a Riemannian manifold.
An edge cost function is defined to search for a desired path in the joint space and reduce the range of motion of the joint.
Second, in this method, by representing concave obstacles in the position space as a combination of non-concave obstacles in the joint space, the distance to the obstacles can be efficiently calculated using a gradient algorithm.
Finally, we validated the proposed approach through experiments and demonstrated that the method can generate trajectories with reduced joint motion.

\section*{Materials and methods}

\subsection*{Surgical robot}
\begin{figure}[t]
\centering
\begin{minipage}[t]{0.49\columnwidth}
    \centering
    \includegraphics[width=1\columnwidth]{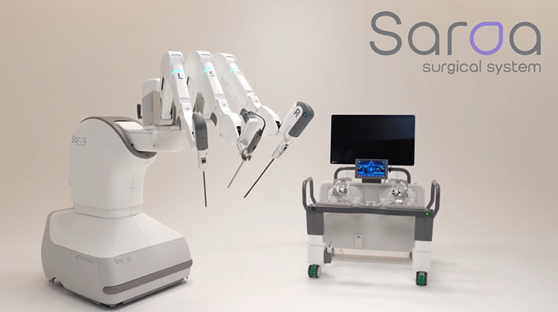}
    \caption{Surgical robot used in this research}
    \label{fig:saroa}
\end{minipage}
\begin{minipage}[t]{0.49\columnwidth}
    \centering
    \includegraphics[width=0.9\columnwidth]{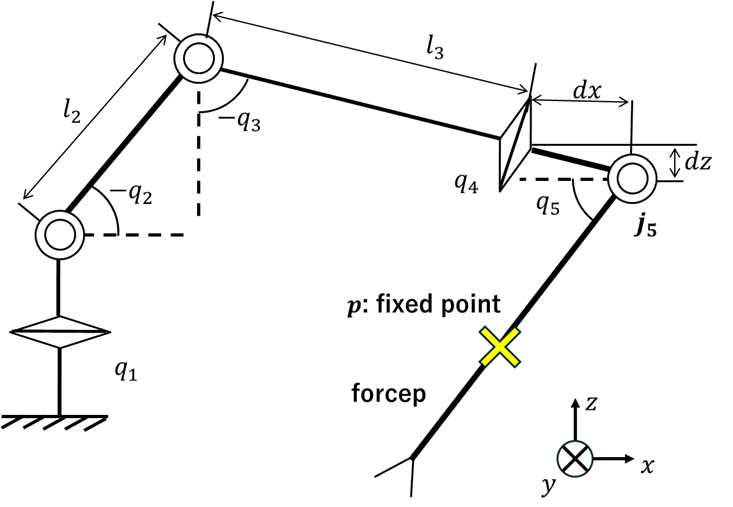}
    \caption{Joint arrangement of the holder robot}
    \label{fig:joint}
\end{minipage}

\end{figure}

Surgical robots for MIS are usually leader-follower systems.
The follower robot consists of forceps manipulators and holder arms. For example, a clinically used robot named Saroa in Japan, the follower robot consists of a robotic forceps and a holder arm to which the forceps are attached (~Fig.\ref{fig:saroa})\cite{Saroa1}$^{,}$\cite{Saroa2}.
The holder arm is a serial type robot with 5-Dofs and a schematic diagram of the arm is shown in  Fig.~\ref{fig:joint}.
Articulated robots generally have six degrees of freedom including tip rotation and can assume any position or posture at the tip.  


In Fig.\ref{fig:joint}, the following symbols are defined.
The joint angle of the $i$ link is defined as $q_i$.
Also, $q_2$, $q_3$, and $q_5$ are given with a minus sign for the calculations of inverse dynamics.
The lengths of the links are shown as $\mathit{l}_2$ and $\mathit{l}_3$, and the position vector of the joint $\mathit{q}_5$ is given as $\mathbf{j}_5$.
The distances on the x- and z-axes of joints 3 and 5 are shown in the figure as $d_x$ and $d_z$, respectively.
The arm has shoulder joints $q_1$ and 
$q_2$ for the yaw and pitch motions at the base. 
The elbow joint $q_3$ is for the yaw motion.
The wrist joints are $q_4$ and $q_5$.
If the horizontal plane is 0 degrees, the range of motion of each joint can be expressed in degrees as follows:
$q_1$ is from -150 to +150,
$q_2$ is from 0 to +90,       
$q_3$ is from -90 to 0,
$q_4$ is from -90 to +90
and $q_5$ is from -14 to +104.
This structure enables the surgical robotic arm not to require the precise alignment of its remote center of mechanism (RCM) with the insertion port\cite{RCM}, to avoid a troublesome setup, and to reduce the forces exerted on the patient's body. 

\subsection*{Joint space obstacle mapping}
\subsubsection*{Riemann metric}
Surgical robot arms with five independent joints constrained by a fixed trocar port, whose configuration is represented by the vector $q_{5d}\in Q\subset\mathbb{R}^5$. Through intraoperative imaging and segmentation, we obtain a forbidden workspace region $X_a\subset\mathbb{R}^3$ that the tool tip must avoid. Our goal is to find a continuous path
$ \gamma : [0,1] \to Q$
such that the path minimizes the Riemannian geodesic length
\begin{equation}
  L(\gamma) = \int_0^1 \sqrt{\dot{\gamma}(t)^\mathsf{T}\,G\bigl(\gamma(t)\bigr)\,\dot{\gamma}(t)}\,\mathrm{d}t
  \label{length}
\end{equation}
where $G(\gamma(t))$ is a composite Riemman metric, while ensuring that at all times the forward-kinematics map $K(q_{5D}(t))$ remains outside the forbidden region, i.e.
$
  \forall t\in[0,1],\quad K\bigl(\gamma(t)\bigr) \notin X_a.
 $
Here, $K:Q\to\mathbb{R}^3$ computes the end-effector position from joint angles.

We chose $q_1$ to $q_3$ because the minimum distance will be the same for any three joints.
We define the following vectors.
\[
\boldsymbol{q}=(q_1,q_2,q_3)^T, 
\hspace{0.2cm}\boldsymbol{q_{5D}}=(q_1,q_2,q_3, q_4, q_5)^T
,\hspace{0.2cm}q_4=f(\boldsymbol{q})
,\hspace{0.2cm}q_5=g(\boldsymbol{q})
\label{forward kine}
\]
Here, $f$ and $g$ denote the forward kinematics equation to calculate the joint angles from $\boldsymbol{q}$.
We can give the following equation from Eq.(\ref{length}).
\begin{equation}
L^2(\boldsymbol{q})=d\boldsymbol{q}^T G_\mathrm{q}(\boldsymbol{q})d\boldsymbol{q}=
\rVert {dq_{5D}}\rVert^2.
\end{equation}
Here, $G_\mathrm{q}(\boldsymbol{q})$ is Riemann metric.
By measuring the length in the space containing the metric, 
we can measure the distance considering the contribution of $q_4$ and $q_5$.
\[
d\boldsymbol{q}=(dq_1, dq_2, dq_3)^T
\]
\[
dq_{5D}=
\begin{pmatrix}
dq_1\\
dq_2\\
dq_3\\
\frac{\partial f}{\partial q_1}dq_1+\frac{\partial f}{\partial q_2}dq_2+\frac{\partial f}{\partial q_3}dq_3\\
\frac{\partial g}{\partial q_1}dq_1+\frac{\partial g}{\partial q_2}dq_2+\frac{\partial g}{\partial q_3}dq_3\\
\end{pmatrix}
=
\begin{pmatrix}
d\boldsymbol{q}\\
\nabla(f)^Td\boldsymbol{q}\\
\nabla(g)^Td\boldsymbol{q}\\
\end{pmatrix}
\]
From the above equations, the right hand side of Eq.(2) can be written as follows.
\[
\rVert {dq_{5D}}\rVert^2 =
d\boldsymbol{q}^Td\boldsymbol{q}+
(\nabla(f)^Td\boldsymbol{q})^2+
(\nabla(g)^Td\boldsymbol{q})^2
\]
Finally, the Riemann metric $G_\mathrm{q}(\boldsymbol{q})$ is given as
\begin{equation}
G_\mathrm{q}(\boldsymbol{q}) = I+\nabla(f(\boldsymbol{q}))^T\nabla(f(\boldsymbol{q}))+
\nabla(g(\boldsymbol{q}))^T\nabla(g(\boldsymbol{q})).
\label{metric}
\end{equation}
From the above equation, we can calculate the shortest path in joint space that includes all five joints.

\subsubsection*{Collision region in joint space}
\begin{figure}[t]
\centering
\begin{minipage}[t]{0.49\columnwidth}
    \centering
    \includegraphics[width=0.8\columnwidth]{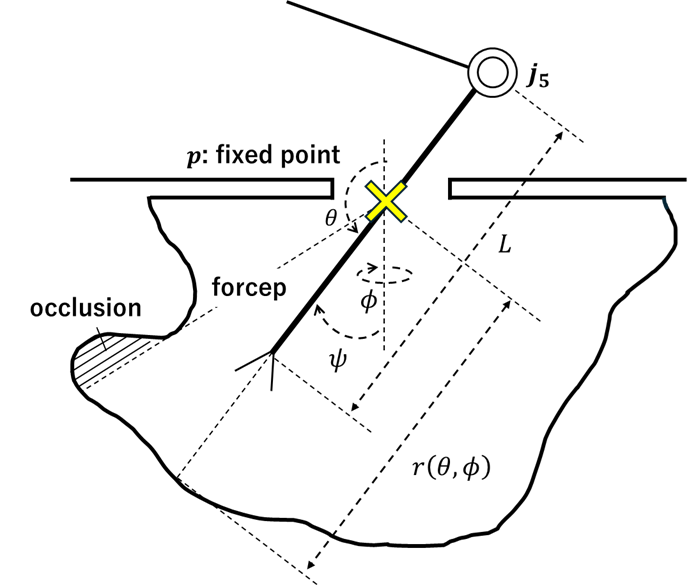}
    \caption{Side view of forceps inserted from pivot point to inner cavity
    \label{fig:cavity}}
\end{minipage}
\begin{minipage}[t]{0.49\columnwidth}
    \centering
    \includegraphics[width=1\columnwidth]
  {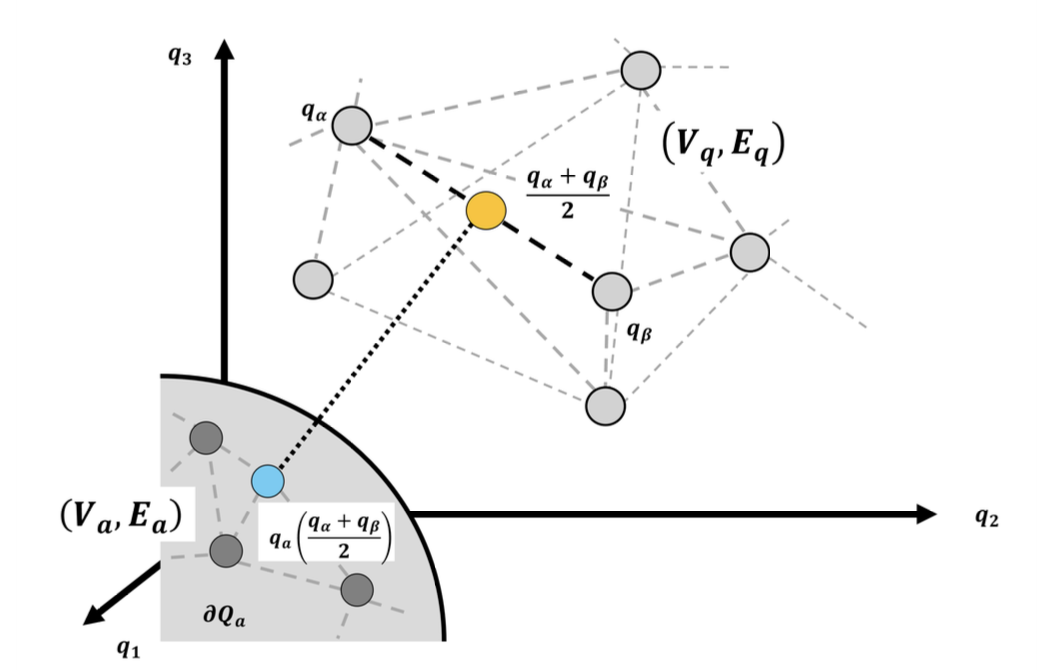}
  \caption{Undirected graphs used in sampling-based shortest path calculation methods}
  \label{fig:undirect1}
\end{minipage}
\end{figure}

Fig.~\ref{fig:cavity} shows the schematic diagram of the forceps in the cavity from the pivot point.
We give polar coordinates around the insertion point as shown in the figure.
The forbidden joint configurations lie on the boundary of the collision region in joint space, denoted $\partial Q_a\subset Q$. We characterize this boundary by the set of all $\boldsymbol{q}$ for which there exists a direction $(\theta,\phi)$ in the half-sphere $[\pi/2,\pi]\times[0,2\pi)$ such that the  
end-effector position satisfies
\[
  x(\boldsymbol{q};p) = p + r(\theta,\phi)\,e_{\theta,\phi},
\]
where $p$ is the trocar port location, $r(\theta,\phi)$ is the segmented radial distance to the first collision along the unit vector. Here, 
\[  e_{\theta,\phi}=\bigl(\sin\theta\cos\phi,\,\sin\theta\sin\phi,\,\cos\theta).
\]
Then, the collision region in joint space is given as 
\[ 
  \partial Q_a=\Bigl\{\bm q\in Q\mid\exists\,(\theta,\phi):x(\bm q;p)=p+r(\theta,\phi)e_{\theta,\phi}\Bigr\}.
\]
We approximate the continuous boundary $\partial Q_a$ by sampling a grid of directions $(\theta_i,\phi_i)$ uniformly over $[\pi/2,\pi]\times[0,2\pi)$ and, for each sample, computing the corresponding Cartesian point $x_i=p+r(\theta_i,\phi_i)e_{\theta_i,\phi_i}$. We then solve the inverse kinematics $q_i=K^{-1}(x_i)$ to obtain discrete boundary configurations in $Q$. Finally, applying Delaunay triangulation to the point set $\{q_i\}$ yields a piecewise-linear mesh that approximates $\partial Q_a$ without explicit enumeration environments.

\subsection*{Joint Space Path Planning}
We define the total metric
\[ 
  G(\boldsymbol{q})=G_{\mathrm{q}}(\boldsymbol{q})+G_{\mathrm{obs}}(\boldsymbol{q}),
\]  
where the kinematic term $G_\mathrm{q}(\boldsymbol{q})$ is given in Eq.(\ref{metric}).
The obstacle term
\[ 
  G_{\mathrm{obs}}(q)=\frac{\sigma}{\|\boldsymbol{q}-\boldsymbol{q}_a\|}\,I,
\]  
with $\boldsymbol{q_a}=\arg\min_{q'\in Q_a}\|\boldsymbol{q}-\boldsymbol{q}'\|$, enforces an infinite cost barrier as the configuration approaches the forbidden boundary $\partial Q_a$.
It is known that the shortest path $\boldsymbol{q}(t)$ in space containing the metric $G(\boldsymbol{q})$ is generally a solution of Lagrange's equation of motion for the following Lagrangian.
\[
L(\boldsymbol{q},\boldsymbol{\dot q})=\frac{1}{2}\boldsymbol{\dot q}^T G(\boldsymbol{q}) \boldsymbol{\dot q},\hspace{0.3cm}
\frac{\partial L}{\partial \boldsymbol{q}}-\frac{d}{dt}
\frac{\partial L}{\partial \boldsymbol{\dot q}}=0
\]
Finally, the following equation is derived (see Appendix A).
\begin{equation}
    G(\bm{q})\ddot{\bm{q}} + \left(\nabla f\dot{\bm{q}}^T \nabla^2 f + \nabla g\dot{\bm{q}}^T \nabla^2 g\right)\dot{\bm{q}}
    - \frac{\sigma_q}{\left\|\bm{q}-\bm{q}_a\right\|^3}\left(\dot{\bm{q}}\dot{\bm{q}}^T - \frac{1}{2}\left\|\dot{\bm{q}}\right\|^2 I\right)\left(I-\frac{\partial\bm{q}_a}{\partial\bm{q}}\right)^T(\bm{q}-\bm{q}_a) = 0\label{eq:kinematics_equation}
\end{equation}
The shortest path in the join space from $\boldsymbol{q}_s$ to $\boldsymbol{q}_t$ at time $T$ is given by solving Eq.(\ref{eq:kinematics_equation}) with the boundary conditions $\boldsymbol{q}(0)=\boldsymbol{q}_s$ and
$\boldsymbol{q}(T)=\boldsymbol{q}_t$.

\subsection*{Proposed edge cost function}
If a convex representation in the joint space is guaranteed, gradient methods achieve fast convergence to the optimal solution.
Gaussian curvature $K$ and mean curvature $H$ of a surface implicitly defined by $F(x, y, z) = 0$ in joint space are given as follows:
\begin{equation}
K=-\frac{\rm{det}\bigg(\begin{matrix}
J^TH(F)J+H_o & J^T(\nabla F) \\
(\nabla F)^T J & 0 \\
\end{matrix}\bigg)}
{J^T|\nabla F|^4},
H=\frac{1}{2|J^T\nabla F|^3}|\nabla F|^T J [\rm{trace}\it(J^T H(F)J-\nabla F+H_o)I)^T-(J^T H(f)J+H_o)]J^T \nabla F. \label{Gaussian}
\end{equation}
\[J=\frac{\partial(x,y,z)}{\partial(q_1,q_2,q_3)},
 (H_o)_{i,j}=\frac{\partial F}{\partial x}\frac{\partial^2 x}{\partial q_i \partial q_j}
+\frac{\partial F}{\partial F}\frac{\partial^2 y}{\partial q_i \partial q_j}
\frac{\partial F}{\partial F}\frac{\partial^2 z}{\partial q_i \partial q_j}
\]
\noindent
The surface is locally non-concave at that point when $K	\geq$0 and $H\geq$0. 
Referring to MRI images, the surfaces of organs can be written as a combination of convex surfaces, while
the walls of body cavities are concave\cite{MRI}.

First, we identify areas where the intracavity shape is represented by a combination of non-concave surfaces within the joint space ($K\geq0$ and $H\geq0$).
Next, we search the path on non-concave surfaces.
Then, we adopt a sampling-based approximation,
since the analytic solution of the boundary-value geodesic problem shown in Eq.(\ref{eq:kinematics_equation}) is intractable under our highly nonlinear metric.
We begin by randomly sampling a finite set of collision-free configurations $V_q\subset Q$, always including the start $\boldsymbol{q}_s$ and the goal $\boldsymbol{q}_t$. Constructing the Delaunay triangulation gives a sparse graph $(V_q,E_q)$ whose edges connect nearby samples.
Each edge $(\boldsymbol{q}_\alpha,\boldsymbol{q}_\beta)\in E_q$ is weighted by an approximation of the path length under $G(\boldsymbol{q})$ plus the barrier penalty. Denoting the midpoint $\boldsymbol{\bar q}=(\boldsymbol{q}_\alpha+\boldsymbol{q}_\beta)/2$ and the closest forbidden configuration $\bm q_a(\bar q)$, the edge cost function is proposed and given by
\begin{equation}
{c(\boldsymbol{q}_\alpha,\boldsymbol{q}_\beta)=} 
   \sqrt{(\boldsymbol{q}_\beta-\boldsymbol{q}_\alpha)^\mathsf{T}
  G(\boldsymbol{\bar q})(\boldsymbol{q}_\beta-\boldsymbol{q}_\alpha)}\,\left[1+\frac{\sigma_q}{\|\ \boldsymbol{\bar q}-\boldsymbol{q}_a( \boldsymbol{\bar q})\|_2}\right].
  \label{edge} \end{equation}
Here,$\sigma_q$ is a setting parameter.
The first term in equation (\ref{edge}) describes the shortest path through the joint space, preventing the robot arm from approaching its operating limits.
The second term of Eq. (\ref{edge}) ensures that edges near obstacles incur a large penalty, discouraging collisions.

A schematic diagram of an undirected graph illustrating path computation is shown in Fig.~\ref{fig:undirect1}.
With all edge costs defined, we run Dijkstra's algorithm on $(V_q,E_q)$ to find the minimum-cost discrete path between the nodes nearest to $\boldsymbol{q}_s$ and $\boldsymbol{q}_t$ \cite{Dij}. Finally, we reinsert the exact endpoints $\boldsymbol{q}_s$ and $\boldsymbol{q}_t$ and apply cubic spline interpolation through the resulting node sequence to generate a smooth continuous trajectory $\gamma(t)$ that approximates the geodesic under $G(\boldsymbol{q})$.


For the spline interpolation, the path is generated using cubic spline interpolation for a sequence of points ($q_0$=$q_s$, $q_1$, $q_2$, ..., $q_{n-1}$, $q_n=q_t$) obtained on a non-directed graph in angle space.


\subsection*{Calculation method}
We compute paths in the region where $\partial Q_a$  can be expressed as a combination of convex surfaces.
Therefore, we partition $\partial Q_a$  into $n$ convex surfaces and partition the graph $(V_a, E_a)$ into $n$ graphs $(V_{a, i}, E_{a, i})(i=1, \dots, n)$ according to this partition.
The point $\bm{q}_{a, i}\in V_{a, i}$ with the minimum distance from $\bm{q}$ in each graph is obtained by the greedy method.
\newpage
The code representing this algorithm is shown in Algorithm \ref{alg:search_q_a}.

\begin{algorithm}
\caption{Algorithm for searching $\bm{q}_a(\bm{q})$}
\begin{algorithmic}[1]
\STATE Construct graph $(V_a, E_a)$ on $\partial Q_a$
\STATE Segment $(V_a, E_a)$ to Convex Patches $(V_{a, i}, E_{a, i})\ (i=1, ..., n)$
\STATE $\bm{q}_a \leftarrow None, d_{min} \leftarrow INF$
\FOR{$i=1$\ to\ $n$}
\STATE $\bm{q}_{a, i} \leftarrow \bm{q}_{init}\in V_{a, i},
d_{min, i} \leftarrow |\bm{q}_{a, i} - \bm{q}|$
\WHILE{true}
\STATE $\bm{q}_{next} \leftarrow None$
\FOR{$\bm{q}_{nei} \leftarrow $neighbor$(\bm{q}_{a, i}, (V_{a, i}, E_{a, i}))$}
\STATE $d\leftarrow|\bm{q}_{nei} - \bm{q}|$
\IF{$d<d_{min, i}$}
\STATE $d_{min, i} \leftarrow d, \bm{q}_{next} \leftarrow \bm{q}_{nei}$
\ENDIF
\ENDFOR
\IF{$\bm{q}_{next} = None$}
\STATE break
\ENDIF
\STATE $\bm{q}_{a, i} \leftarrow \bm{q}_{next}$
\ENDWHILE
\IF{$d_{min, i} < d_{min}$}
\STATE $\bm{q}_a \leftarrow \bm{q}_{a, i}, 
d_{min} \leftarrow d_{min, i}$
\ENDIF
\ENDFOR
\STATE return $\bm{q}_a$
\end{algorithmic}
\label{alg:search_q_a}
\end{algorithm}
\noindent
Since $\bm{x}_a$ is concave, it is not possible to divide $\bm{q}_a$ into convex surfaces and obtain each of them by using the greedy method as in the case of $\bm{q}_a$. Therefore, it is obtained by searching all the sampled points on $\partial X_a$.
The number of nodes was set to $n$=5000 from the pre-examination.

Consider a path planning method in position space for comparison.
Let $E_x$ be the edge stretched by the 3-dimensional Delaunay partition of a randomly sampled point $V_x$ in the region $X$ and define an undirected graph $(V_x, E_x)$.
For each edge $(\bm{x}_\alpha, \bm{x}_\beta)\in E_x$, define the edge cost as follows.
Here, $\bm{x}_a$ represents the closest point to $\bm{x}$ on $\partial X_a$, which is obtained as a point in a node on the polygon as in $q$ space.
We find the closest $\bm{x}_{s, x}, \bm{x}_{t, x}\in V_x$ to $\bm{x}_s, \bm{x}_t$ and use Dijkstra method to find the shortest path between these two points $(\bm{x}_1 = \bm{x}_{s, x}, \bm{x}_2, \dots, \bm {x}_n=\bm{x}_{t, x})$.
The curve $C_{x, opt}$ obtained by applying cubic spline interpolation is the approximate shortest path for the point sequence
$ (\bm{x}_s, \bm{x}_1, \bm{x}_2, \dots, \bm{x}_n, \bm{x}_t).
$
\begin{equation}
    c\left(\bm{x}_\alpha, \bm{x}_\beta\right) = \left\|\bm{x}_\alpha - \bm{x}_\beta\right\|
    \left(1+\frac{\sigma_x}{\left\|\frac{\bm{x}_\alpha + \bm{x}_\beta}{2} - \bm{x}_a\left(\frac{\bm{x}_\alpha + \bm{x}_\beta}{2}\right)\right\|}\right)
    \label{cost_x}
    \end{equation}
Here,$\sigma_x$ is a setting parameter.

\subsection*{Mimic organs}
\begin{figure}[t]
    \centering
    \includegraphics[width=0.7\columnwidth]{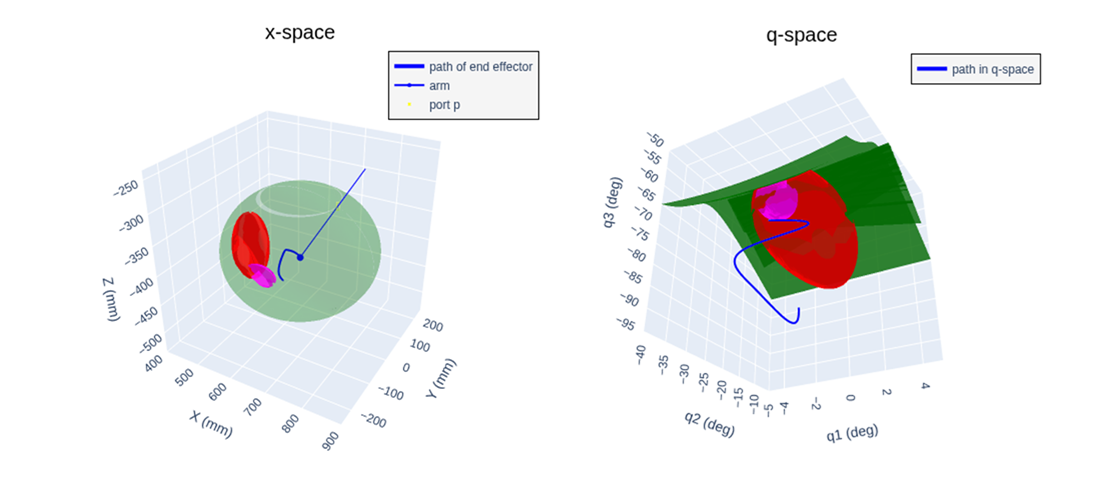}
\caption {Transferring the mimic organ of cholecystectomy from position space (left) to joint space (right)} 
\label{fig:chol}
\end{figure}

Well-known robotic surgeries, cholecystectomy and prostatectomy, were simulated with mimic organs.
To simulate a cholecystectomy, a sphere representing the gallbladder is placed within the body cavity of a concave ellipsoid. The ellipoid is give as follows.
\[
  \partial X_a=
  \left(
  \boldsymbol{p}+c+\left(
  \begin{matrix}
  a \cos\theta \cos\phi\\
  a \sin\theta \cos\phi\\
  b \cos\theta
\end{matrix}
\right)\ 0< \theta<\pi,
  0<\phi<2\pi 
  \right) 
  \]
Here, $c=[-100, 0, 50\sqrt3]$, $a$=200 and $b$=100
are selected. The units of $a,b$ and $c$ are mm. 
The ellipsoidal mimic liver is located at the center point $[500, 0, -320-50\sqrt3]$ and has a radius of $[50, 60, 60]$.
The ellipsoidal mimic gallbladder is located at the center point $[520, 0, -370-50\sqrt3]$ and has a radius of $[50, 20, 20]$.
Its shape is shown in the left diagram of  Fig.\ref{fig:chol}. 
The red part is the mimic lever and the pink part is the mimic gallbladder.

\begin{figure}[t]
    \centering
    \includegraphics[width=0.5\columnwidth]{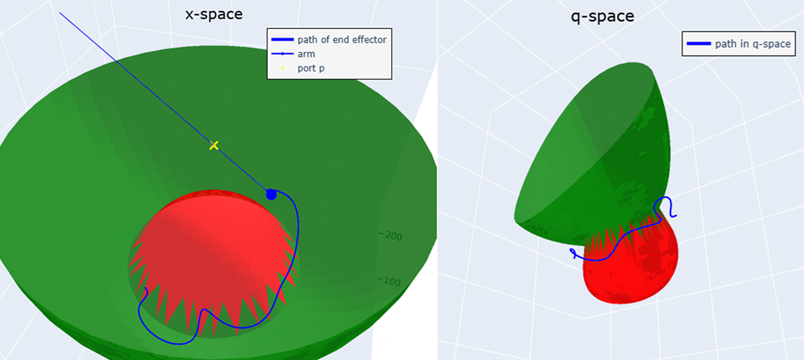}
\caption {Transferring the mimic organ of bladder from position space (left) to joint space (right)} 
\label{fig:pro}
\end{figure}

The bladder in a prostatectomy was simulated using a hemisphere representing the bladder placed in a concave body cavity as shown in the left diagram of Fig.\ref{fig:pro}.
We set the case where the approximate shape of the avoidance region in the body cavity $x_a$ is hemispherical.
A sphere of center (0, 0, 0.3$L$) and radius of 0.15$L$ is placed on a hemispherical surface with surface area $R=$0.3$L$.
\[
  \partial X_a=\{\boldsymbol{p}+d\boldsymbol{e}_z+R e_{\theta.\phi}|\arccos\left(-\frac{d}{R}\right)\ < \theta<\pi,
  0<\phi<2\pi\} 
  \]
Here, $R$ is the radius of the hemispherical and $L=$300mm is the length of the robotic forceps.
The red part is the mimic bladder.

\subsection*{Evaluation of the proposed method}
We compare the path planning method in position space and the proposed method in joint space.
In comparing planned paths in two spaces, we need to determine $\sigma_q$ in Eq.(\ref{edge}) and $\sigma_x$ in Eq.(\ref{cost_x}) for the equivalent comparison.
The first scenario in which the proposed method is used is to guide the forceps from near the insertion point to near the target organ.
In this scenario, a 5.0 mm margin to the organ allows for adequate guidance.
Therefore, we set $\sigma_q$  as 0.842 deg which is equal in joint space for $\sigma_x=$5.0 mm.

$\partial X_a$ is obtained as mesh data. From this mesh, one triangle $(\bm{x}_1, \bm{x}_2, \bm{x}_3)$ is taken. For the unit normal vector $\bm{n}$ of this triangle, determine $\bm{x}_+, \bm{x}_-$ as follows.
\[
    \bm{x}_\pm = \frac{1}{3}(\bm{x}_1 + \bm{x}_2 + \bm{x}_3) \pm \frac{\sigma_x}{2}\bm{n}\ 
\]
Applying inverse kinematics to this $\bm{x}_+, \bm{x}_-$ to calculate $\bm{q}_+ = K^{-1}(\bm{x}_+), \bm{q}_- = K^{-1}(\bm{x}_-)$ and determine $\hat{\sigma}_q$ as $
    \hat{\sigma}_q=\left\|\bm{q}_+ - \bm{q}_- \right\|.
$
This is computed for all triangles in the mesh and averaged to obtain $\sigma_q$. That is, if the number of meshes is $M$, we obtain the following equation.
\[
    \sigma_q = \frac{1}{M}\sum_{(\bm{x}_1, \bm{x}_2, \bm{x}_3) \in \partial X_a} \hat{\sigma}_q
\]
$\hat{\sigma}_q$ calculated for each mesh represents the amount of movement in the joint space at each point on $\partial X_a$ when moved by $\sigma_x$ perpendicular to $\partial X_a$.
By taking this average value, we can set up a buffer in the joint space that is equivalent to the buffer set up in the position space.
For the path generation, it is necessary to evaluate routes $C_{x, opt}$ in the position space and 
$C_{q, opt}$ in the joint  space with a common index. 
This time $C_{q, opt}$ is converted to the position space and evaluated.

We evaluated the path generation of the proposed method in the joint space with the following indices.
One is the insertion angle of the forceps. 
We define the angle $\psi$ as
\begin{equation}
    \psi(\bm{x}) = -\arctan\left(\frac{\sqrt{(x-p_x)^2 + (y-p_y)^2}}{z-p_z}\right)
    \label{eq:ang}
\end{equation}
which is the angle between the forceps and the vertical axis in Fig.~3.
The range of the angle is $0 < \psi(\bm{x}) < \frac{\pi}{2}$.
The larger this $\psi(\bm{x})$ is, the more horizontally the forceps is inserted against the body wall.
Because the body wall is thick, inserting forceps horizontally causes torsion at the port.
Therefore, this $\psi( \bm{x})$ should be small in the path of generation. 
The mean $\psi_{ave}(C)$ and maximum values $\psi_{max}(C)$ of the angles were observed.
\begin{equation}
    \psi_{ave}(C) = \frac{1}{s_C}\int_0^{s_C} \psi\left(\bm{x}(s)\right) ds, 
    \psi_{max}(C) = \max_{\bm{x}\in C}\psi(\bm{x})
\end{equation}

The path length is also used as the index.
The root mean square of the derivative with respect to the path length was calculated as follows.
\begin{equation}
    \psi'_{RMS}(C) = \sqrt{\frac{1}{s_C}\int_0^{s_C}\left\{\frac{d\psi(\bm{x}(s))}{ds}\right\}^2ds}
\end{equation}
Here, $s_C$ is the path length and $s$ is the arc length of the path.

\section*{Results}
\subsection*{Transfer to Riemannina manifold}
The result of transferring the mimic organ of cholecystectomy from position space to joint space is shown in Fig~\ref{fig:chol}.
The left figure shows the shape in position space, and the right figure shows the shape  transferred to joint space.
The convexity of the shape was evaluated using Eq.(\ref{Gaussian}). 
The mimic lever and gallbladder are concave in joint space as shown in the right diagram of Fig~\ref{fig:chol}.
As mentioned earlier, the first scenario is to guide the robotic forceps from the insert point to the starting point of the task. 
The estimated path is shown by the blue line in the figure. 
We confirmed that the  path lies in the non-concave region.
Fig.~\ref{fig:pro} shows the mimic bladder transferred from position space to joint space.
It can be seen that the shape that is concave in position space becomes convex in joint space.

For surfaces composed of multiple convex surfaces, the shortest distance in joint space can be obtained by applying the gradient descent method to each convex surface independently and taking the minimum of the output values, which is more efficient than a full search.
Thus, the fact that $\partial Q_a$ consists of a combination of convex surfaces is a desirable condition for path planning, and this problem setting, where $\partial X_a$ includes concave regions, has the advantage of path finding in the joint space.

\subsection*{Experimental results}
The effectiveness of the proposed method was also confirmed by experiments.
The parameters used in the experiments are as follows. 
   $|V_q|, |V_x| = 1000$, 
        $|V_a| = 5000$, 
        $\sigma_x = 5.0 \rm{mm} $,
        $N = 30$,         $L = 500 \rm{mm}$ and
        $k(=R/L)= 0.5, p =$ [750, 0, -300].
The start point and end points were set as
[705,-26, -330] and [656 -26, -378].

\begin{figure}[h]
\centering
    \centering
    \includegraphics[width=0.5\columnwidth]  {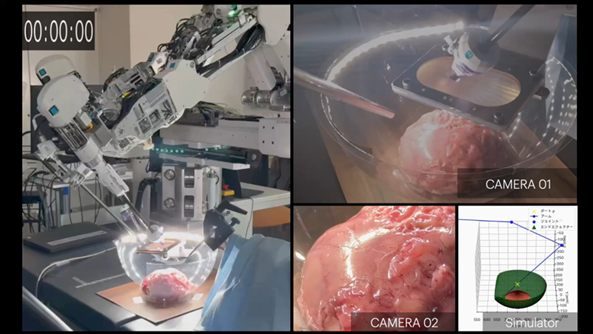}
  \caption {Experimental scenery}
  \label{fig:exp0}
  \vspace{-0.5cm}
\end{figure}

\begin{figure}[h]
\centering
  \begin{minipage}[t]{0.48\columnwidth}
    \centering
      \includegraphics[width=0.8\textwidth]
{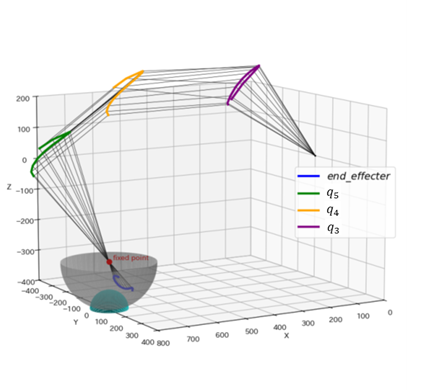}
  \end{minipage}
  \begin{minipage}[t]{0.48\columnwidth}
    \centering
    \includegraphics[width=0.8\textwidth]
{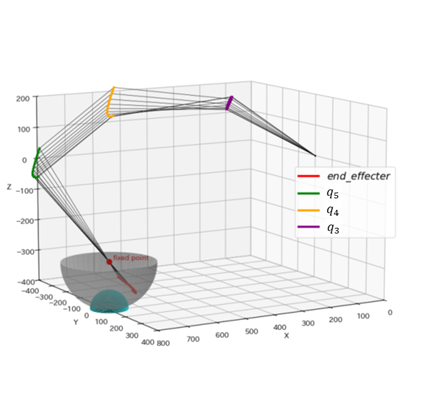}
   \end{minipage}
   \vspace{-0.5cm}
    \caption{Movement profile from the start point to end point searched in position space (left) and in joint space (right: proposed method)}
  \label{fig:motion}
\end{figure}

Fig.~\ref{fig:exp0} shows the scenery during experiment.
We used the mimic organ of bladder.
Fig.~\ref{fig:motion} shows the movement profile of the robot arm from the start point to end point.
It can be seen that the proposed method results in smaller angular movements.

 Table 1 shows the calculation results of the indicators
shown in Eq.(8),(9), and (10). 
The proposed method has smaller index values than the path planning in position space.
Experimental result demonstrated that the method can generate trajectories with reduced joint motion.


\begin{table}[b]
    \centering
    \caption{Index values calculated from experimental results}
    \begin{tabular}{c|c|c|c}\hline
        &$\psi_{ave} $[deg] & $\psi_{max} $[deg] & $\dot\psi_{max} $[deg/mm] \\\hline
        Proposed method & 31.8 & 40.3 & 0.214  \\\hline
        Path planning in position space &  39.5 & 47.7 & 0.256 \\\hline
          \end{tabular}
    \label{tab:params2}
\end{table}

\section*{Conclusion}
In this paper, we proposed a path finding method on Riemannian manifolds in joint space for a surgical robot for MIS.
We confirmed that the mimic organs were represented by a combination of non-concave surfaces in the joint space of the robot arm.
The proposed method achieved faster computation by using the greedy method in the joint angle space to calculate the path avoiding collision to the organs.
We validated the proposed approach through experiments in anatomically informed abdominal environments and demonstrated that our planner yields trajectories with reduced joint motion and port stress. 
These results highlight the potential of our framework as a core component in semi‐autonomous surgical systems that can be implemented using existing clinical robotic platforms.

Future works are as follows.
For more complex obstacles, the necessary conditions for having a convex representation in joint space need to be clarified.
For the generated trajectories, we plan to consider collision avoidance methods other than spline interpolation used in this study.
Consideration of avoiding interference between robot arms and application to actual intracorporeal environments with more complexed shapes.


\newpage
\begin{thebibliography}{9}

\bibitem{MIS}
K.H.Fuchs, Minimally Invasive Surgery,
Endoscopy, 34(2): 154-159, 2022


\bibitem{Mare}
M. Diana and J. Marescaux, Robotic surgery, Journal of British Surgeru, 102(2): e15-e28, 2015 

\bibitem{Emer}
V. Vitiello, S. L. Lee, T. P. Cundy, G-Z Yang,
Emerging Robotic Platforms for
Minimally Invasive Surgery,
IEEE Revies in Biomedical Eng.,6, 111-125, 2012

\bibitem{Kawa}
K.Kawashima, T.Kanno, T.Tadano, Robots in laparoscopic surgery: current and future status,BMC Biomedical Engineering, 1-6, 2019

\bibitem{Cepo}
F. Cepolina, R. P. Razzoli,
An introductory review of robotically assisted surgical
systems, Int J Med Robot. 18(4):e2409, 2022

\bibitem{kawa1}
K. Watanabe, T. Kanno, K. Ito, K. Kawashima, Single Master Dual Slave Surgical Robot with Automated Relay of Suture Needle, IEEE Transactions on Industrial Electronics, 65(8): 6343-6351, 2018

\bibitem{Tam}
T. Haidegger, Autonomy for Surgical Robots: Concepts and Paradigms,
IEEE Transactions on Medical Robotics and Bionics, 1(2):65-76,2019 

\bibitem{Han}
J.Han, J.Davids, H.Ashrafian, A.Darzi, D.S.Elson, M.Sodergren,
A systematic review of robotic surgery: From supervised paradigms to fully autonomous robotic approaches,The International Journal of Medical Robotics and Computer Assisted Surgery, 18(2):1-11, 2022



\bibitem{PRM}
L. E. Kavraki, P. Svestka, J.-C. Latombe and M. H. Overmars, Probabilistic roadmaps for path planning in high-dimensional configuration spaces, IEEE Trans. Robot. Automat., 12(4):566-580, 1996

\bibitem{PRM1}
H. Akbaripour, E. Masehian,
Semi-lazy probabilistic roadmap: a parameter-tuned, resilient and robust path planning method for manipulator robots,
The International Journal of Advanced Manufacturing Technology,
89:1401–1430, 2017

\bibitem{PRM2}
X. Tang, W. Shang, Manifold Approximation-Based Probabilistic
Roadmap Approach for Constrained Path
Planning of Mobile Manipulators, IEEE/ASME Transactions on Mechatronics, 2025 

\bibitem{PRM3}
S. Chen et al., Improved Path Planning and Controller
Design Based on PRM, IEEE Access, 12, 2025 

\bibitem{RRT}
S. LaValle, Rapidly-exploring random trees:A new tool for path planning, Research Report 9811, 1998


\bibitem{kara}
S. Karaman and E. Frazzoli, Incremental sampling-based algorithms for optimal
motion planning, The international journal of robotics research,
30(7): 846–894, 2011.

\bibitem{Wei}
K. Wei and B. Ren, A method on dynamic path planning for robotic
manipulator autonomous obstacle avoidance based on an improved rrt
algorithm, Sensors, 18(2):571, 2018

\bibitem{RRT2}
A. Shaarawya, A. Rastegarpanaha, R. Stolkin,
Task-aware motion planning in constrained environments using
GMM-informed RRT planners, Robotics and Computer-Integrated Manufacturing, 97, 2026

\bibitem{DRL}
B. Zhao, Y. Wu, C. Wu, R. Sun,
Deep reinforcement learning
trajectory planning for robotic
manipulator based on simulation efficient
training,
Scientific Reports,15:8286, 2025

\bibitem{DRL2}
L-S. Schneider, J. Peng, A. Maier,
Robot movement planning
for obstacle avoidance using
reinforcement learning,
Scientific Reports, 15:32506, 2025



\bibitem{Li}
Y.Li, R. Cui, Z. Li, Neural network approximation based nearoptimal
motion planning with kinodynamic constraints using RRT.
IEEE Trans Ind Electron 65(11):8718–8729, 2018


\bibitem{Meng}
B. H. Meng, I.S.Godage, I. Kanj,
RRT*-Based Path Planning for Continuum Arms,
IEEE Robotics and Automation Letters, 7(3), 6830-6836, 2022

\bibitem{Kang}
M. Kang, Q.Chen, Z. Fan, C. Yu, Y. Wang, X. Yu,
A RRT based path planning scheme for multi-DOF robots in unstructured environments, 
Computers and Electronics in Agriculture, 218,108708, 2024

\bibitem{Chair}
I. Chairez et.al.,
Finite-Time Output Robust Control for Restricted Joint Flight Emulator Robotic Arm With Adaptive Tangent Barrier Gains,
IEEE Access, 13, 23700-23716, 2025 

\bibitem{BF2}
M. Yu et al.,
Efficient Motion Planning for Manipulators with
Control Barrier Function-Induced Neural Controller, IEEE ICRA, 2024


\bibitem{Klein}
H. Klein, N. Jaquire, A. Meixner, T. Asfour, 
On the design of region-avoiding in metrics for collision-safe motion generation on Riemaannian manifolds, IEEE IROS, 2346-2353,2023

\bibitem{MRI}
F.M. Sanchez-Margallo, et al., Anatomical changes due to pneumoperitoneum analyzed by mri: an experimental study in pigs. Surgical and radiologic anatomy, 2011



\bibitem{SDF1}
N. Ratliff et al.,
CHOMP: Gradient Optimization Techniques for Efficient Motion Planning, IEEE ICRA, 2009


\bibitem{SDF2}
T. Zhang et al.,
Continuous Implicit SDF Based
Any-shape Robot Trajectory Optimization, IROS, 2023

\bibitem{Rie}
N. Jaquier, L. Rozo, S. Calinon, M. Bürger,
Bayesian Optimization Meets Riemannian Manifolds in Robot Learning,
3rd Conference on Robot Learning (CoRL), 2019

\bibitem{Saroa1}
K Iwatani et al.,
Initial experience of a novel surgical assist robot “Saroa” featuring tactile feedback and a roll-clutch system in radical prostatectomy,
Scientific Reports, 14, 2024 

\bibitem{Saroa2}
H. Nakanishi et al.,
In vivo evaluation of tissue damage from varying grasping forces using the Saroa surgical system,
Scientific Reports, 15, 2025

\bibitem{RCM}
S. Yoshida, T. Kanno, K. Kawashima, Surgical Robot with Variable Remote Center of Motion Mechanism Using Flexible Structure, ASME Journal of Mechanism and Robotics, 10(3), 2018

\bibitem{Dij}
X. Wang et al.,
Path planning of scenic spots based
on improved A* algorithm,
Scientific reports, 12:1320, 2022




\end{thebibliography}

\section*{Funding declaration}
This work was supported in part by KAKEN 25H00717.

\section*{Author contributions statement}
Y.Y., M.S. and S.H. proposed the method.
Y.Y., M.S., S.H. and K.K. conceived the experiments, 
Y.Y. and T.K. conducted the calculations,
T.K. and T.M. conducted the experiments, Y.Y.,
M.S., T.K., T.M. and K.K. analysed the results.  All authors reviewed the manuscript. 

\section*{Additional information}

\section*{Appendix A}
Eq.(\ref{eq:kinematics_equation}) is calculated as follows.
\[
    L(\bm{q}, \dot{\bm{q}}) = \frac{1}{2}\dot{\bm{q}}^T G_q(\bm{q}) \dot{\bm{q}} + \frac{1}{2}\dot{\bm{q}}^T G_{\mathrm{obs}}(\bm{q}) \dot{\bm{q}}
    \]
Since Lagrange's equation of motion is linear with respect to the Lagrangian, $G(\bm q)$ is derived by adding the two Lagrangian equations of motion.
\[
    L_q(\bm{q}, \dot{\bm{q}}) = \frac{1}{2}\dot{\bm{q}}^T G_q(\bm{q}) \dot{\bm{q}}, 
       L_{\mathrm{obs}}(\bm{q}, \dot{\bm{q}}) = \frac{1}{2}\dot{\bm{q}}^T G_{\mathrm{obs}}(\bm{q}) \dot{\bm{q}}
\]
First, we calculate Lagrangian $L_q(\bm{q}, \dot{\bm{q}})$.
The equation can be written as follows.
\[
    L_q(\bm{q}, \dot{\bm{q}}) = \frac{1}{2}\left(\left\|\dot{\bm{q}}\right\|^2 + \left(\nabla f^T \dot{\bm{q}}\right)^2 + \left(\nabla g^T \dot{\bm{q}}\right)^2 \right)
\]
Hereby, the gradient $\nabla f(\bm q)$ and Hessian $\nabla^2 f(\bm q)$ is written as $\nabla f$ and $\nabla^2 f$ omitting the argument $\bm q$. The following equations can be calculated.
\[
    \frac{\partial L_q}{\partial \bm{q}} = \nabla^2 f\dot{\bm{q}}\nabla f^T\dot{\bm{q}} + \nabla^2 g\dot{\bm{q}}\nabla g^T\dot{\bm{q}}, \hspace{0.3cm}
    \frac{d}{dt}\frac{\partial L_q}{\partial \dot{\bm{q}}} = \frac{d}{dt}\left(G_q(\bm{q})\dot{\bm{q}}\right)\nonumber
    =G_q(\bm{q})\ddot{\bm{q}} + \frac{dG_q(\bm{q})}{dt}\dot{\bm{q}}
\]
\[
    \frac{dG_q(\bm{q})}{dt} = \frac{d}{dt}\left(I+\nabla f \nabla f^T + \nabla g \nabla g^T\right)\nonumber
    \]
    \[
    = \frac{d\left(\nabla f\right)}{dt}\nabla f^T +
    \nabla f \left(\frac{d\left(\nabla f\right)}{dt}\right)^T \\ + \frac{d\left(\nabla g\right)}{dt}\nabla g^T + \nabla g \left(\frac{d\left(\nabla g\right)}{dt}\right)^T\nonumber
    = \nabla^2 f\dot{\bm{q}}\nabla f^T + \nabla f\dot{\bm{q}}^T\nabla^2 f 
    \\ +\nabla^2 g\dot{\bm{q}}\nabla g^T + \nabla g\dot{\bm{q}}^T\nabla^2 g
  \]
As a result, equation of motion for Lagrangian $L_q$ is given as 
\[
    G_q\left(\bm{q}\right)\ddot{\bm{q}} + \nabla f\dot{\bm{q}}^T \nabla^2 f\dot{\bm{q}} + \nabla g\dot{\bm{q}}^T \nabla^2 g\dot{\bm{q}} = 0.
\]
Next, we calculate the equation of motion for Lagrangian $L_{\mathrm{obs}}$.
\[
    L_{\mathrm{obs}}(\bm{q}, \dot{\bm{q}}) = \frac{\sigma_q \left\|\dot{\bm{q}}\right\|^2}{2\left\|\bm{q_a}-\bm{q}\right\|}
\]
The following equation is calculated.
\[
    \frac{\partial L_{\mathrm{obs}}}{\partial \bm{q}} = -\frac{\sigma_q \left\|\dot{\bm{q}}\right\|^2}{2\left\|\bm{q}-\bm{q}_a\right\|^3}\left(I-\frac{\partial \bm{q}_a}{\partial \bm{q}}\right)^T\left(\bm{q}-\bm{q}_a\right),\\
    \frac{d}{dt}\frac{\partial L_{\mathrm{obs}}}{\partial \dot{\bm{q}}} = \frac{\sigma_q}{\left\|\bm{q}-\bm{q}_a\right\|}\ddot{\bm{q}}-\sigma_q\frac{\dot{\bm{q}}^T\left(I - \frac{\partial \bm{q}_a }{\partial \bm{q}}\right)^T\left(\bm{q}-\bm{q}_a\right)}{\left\|\bm{q}-\bm{q}_a\right\|^3}\dot{\bm{q}}
\]
Then, the equation of motion is give as follows:
\[
  \frac{\sigma_q}{\left\|\bm{q}_a - \bm{q} \right\|} \ddot{\bm{q}} 
  - \frac{\sigma_q}{\left\| \bm{q}_a - \bm{q} \right\|^3} \cdot
  \Biggl\{
    \dot{\bm{q}}^{\mathsf{T}} \left(I - \frac{\partial \bm{q}_a}{\partial \bm{q}} \right)^{\mathsf{T}} (\bm{q}_a - \bm{q}) \, \dot{\bm{q}} 
     - \frac{1}{2} \left\| \dot{\bm{q}} \right\|^2 \left(I - \frac{\partial \bm{q}_a}{\partial \bm{q}} \right)^{\mathsf{T}} (\bm{q}_a - \bm{q})
  \Biggr\}
  = 0.
\]
The second term in the above equation 
$\dot{\bm{q}}^T\left(I-\frac{\partial\bm{q}_a}{\partial\bm{q}}\right)^T(\bm{q}-\bm{q}_a)$
is a scalar. 
Therefore, by transforming the above equation, we obtain the following equation.
\[
   \dot{\bm{q}}^T\left(I-\frac{\partial\bm{q}_a}{\partial\bm{q}}\right)^T(\bm{q}-\bm{q}_a)\ \dot{\bm{q}} - \frac{1}{2}\left\|\dot{\bm{q}}\right\|^2\left(I-\frac{\partial\bm{q}_a}{\partial\bm{q}}\right)^T(\bm{q}-\bm{q}_a)\nonumber\\
    =\ \dot{\bm{q}}\dot{\bm{q}}^T\left(I-\frac{\partial\bm{q}_a}{\partial\bm{q}}\right)^T(\bm{q}-\bm{q}) - \frac{1}{2}\left\|\dot{\bm{q}}\right\|^2\left(I-\frac{\partial\bm{q}_a}{\partial\bm{q}}\right)^T(\bm{q}-\bm{q}_a)\nonumber\\\]
    \[
    =\left(\dot{\bm{q}}\dot{\bm{q}}^T - \frac{1}{2}\left\|\dot{\bm{q}}\right\|^2 I\right)\left(I-\frac{\partial\bm{q}_a}{\partial\bm{q}}\right)^T(\bm{q}-\bm{q}_a).
\]
The equation can be organized as follows.
\[
    \frac{\sigma_q}{\left\|\bm{q}-\bm{q}_a\right\|}\ddot{\bm{q}} - \frac{\sigma_q}{\left\|\bm{q}-\bm{q}_a\right\|^3}\left(\dot{\bm{q}}\dot{\bm{q}}^T - \frac{1}{2}\left\|\dot{\bm{q}}\right\|^2 I \right)
     \left(I-\frac{\partial\bm{q}_a}{\partial\bm{q}}\right)^T(\bm{q}-\bm{q}_a) = 0\label{eq:a-kinematics}
\]
Finally, the following equation is derived.
\[
    G(\bm{q})\ddot{\bm{q}} + \left(\nabla f\dot{\bm{q}}^T \nabla^2 f + \nabla g\dot{\bm{q}}^T \nabla^2 g\right)\dot{\bm{q}} \nonumber
    - \frac{\sigma_q}{\left\|\bm{q}-\bm{q}_a\right\|^3}\left(\dot{\bm{q}}\dot{\bm{q}}^T - \frac{1}{2}\left\|\dot{\bm{q}}\right\|^2 I\right) \left(I-\frac{\partial\bm{q}_a}{\partial\bm{q}}\right)^T(\bm{q}-\bm{q}_a) = 0\label{eq:kinematics_equation}
\]

\end{document}